\documentclass[letterpaper, 10 pt, conference]{ieeeconf}  

\IEEEoverridecommandlockouts                              

\usepackage{amsmath,amsfonts}
\usepackage{algorithmic}
\usepackage{algorithm}
\usepackage{array}
\usepackage[caption=false,font=normalsize,labelfont=sf,textfont=sf]{subfig}
\usepackage{textcomp}
\usepackage{stfloats}
\usepackage{url}
\usepackage{verbatim}
\usepackage{graphicx}
\usepackage{cite}
\usepackage{hyperref}
\usepackage{xfrac}
\usepackage{bm}
\usepackage{makecell}
\usepackage{xcolor}

\title{\LARGE \bf
From Ceilings to Walls: Universal Dynamic Perching of Small Aerial Robots on Surfaces with Variable Orientations  }

\author{Bryan Habas$^{1}$, Aaron Brown$^{2}$, Donghyeon Lee$^{1}$, Mitchell Goldman$^{1}$, 
Bo Cheng$^{1}$, \textit{IEEE, Member}
\thanks{$^{1}$Biological and Robotic Intelligent Fluid Locomotion Lab, Department of Mechanical Engineering, The Pennsylvania State University, University Park, PA 16802, USA. Corresponding to B.C. {\tt\small buc10@psu.edu}}%
\thanks{$^{2}$Air Vehicle Intelligence and Autonomy Lab, Department of Aerospace Engineering, The Pennsylvania State University, University Park, PA 16802, USA}
}

\begin{document}

\maketitle
\pagestyle{empty}
\pagestyle{empty}

\begin{abstract}

This work demonstrates universal dynamic perching capabilities for quadrotors of various sizes and on surfaces with different orientations. By employing a non-dimensionalization framework and deep reinforcement learning, we systematically assessed how robot size and surface orientation affect landing capabilities. We hypothesized that maintaining geometric proportions across different robot scales ensures consistent perching behavior, which was validated in both simulation and experimental tests. Additionally, we investigated the effects of joint stiffness and damping in the landing gear on perching behaviors and performance. While joint stiffness had minimal impact, joint damping ratios influenced landing success under vertical approaching conditions. The study also identified a critical velocity threshold necessary for successful perching, determined by the robot's maneuverability and leg geometry. Overall, this research advances robotic perching capabilities, offering insights into the role of mechanical design and scaling effects, and lays the groundwork for future drone autonomy and operational efficiency in unstructured environments.

\end{abstract}

\section*{SUPPLEMENTARY MATERIAL}
\noindent \textbf{Video:} \url{https://youtu.be/llRTp3dR5DE}

\section{Introduction}
\noindent Dynamic perching is a routine feat achieved by a diverse array of biological fliers \cite{lee1993visual, carruthers2010mechanics,srinivasan2000honeybees,liu2019flies,riskin2009bats}. These fliers are often capable of robust perching on surfaces regardless of orientation, allowing them to land in unstructured environments to scout territory, evade predators, or rest. In recent years, perching has emerged as a key area of interest in aerial robotics, crucial for not merely emulating nature, but for enhancing robot autonomy in unstructured settings \cite{zufferey2008bio}. Perching enables flying robots to perform tasks such as visual inspection\cite{seo2018drone,irizarry2012usability}, surveillance \cite{mishra2020drone, kim2018drone}, and reconnaissance more efficiently, eliminating the need for sustained flight. Presently, battery constraints limit these robots to the life of tens of minutes of active operation \cite{cesare2015multi}, underscoring the value of perching in extending operational life by reducing unnecessary energy expenditure. This is particularly relevant in urban landscapes abundant with perching opportunities such as walls, ledges, and ceilings, where robots can execute more diverse tasks, from passive monitoring to serving as communication relays in emergency environments \cite{khalil2019using}.

Previous work by the authors have demonstrated that a highly maneuverable nano-scale quadrotor, approximately 3.25" in diameter, can successfully land on inverted surfaces (i.e., ceilings) \cite{habas2022optimal,habas2022deep,habas2024flies}, while also exploring the effects of leg geometry on inverted landing capabilities in similarly sized quadrotors \cite{habas2024flies}. This was achieved using either a two-stage machine learning framework (combining supervised and unsupervised learning) \cite{habas2024flies} or a deep reinforcement learning framework \cite{habas2022deep}. However, several key questions remain to be addressed: 1) How do landing surface orientations affect perching capabilities and can we extend our previous learning frameworks to achieve universal perching capabilities on surfaces of variable orientations (i.e., omnidirectional landing)? 2) How do system maneuverability, leg geometry, and material properties such as hinge stiffness and damping influence landings? Additionally, how do these factors scale across varying robot sizes?

\begin{figure}[!t]
    \centering
    \includegraphics[width=1.0\columnwidth]{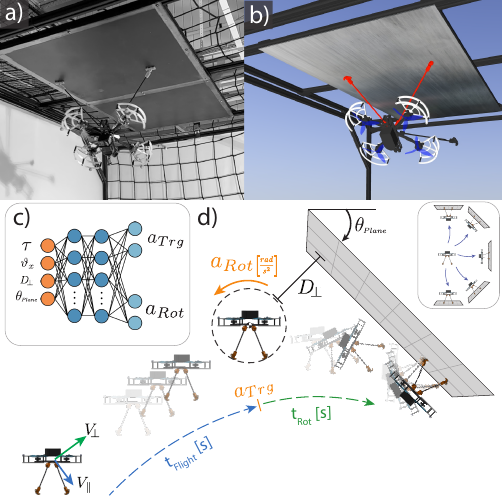}
    \caption{(a) Image capturing the body-swing stage of a quadrotor during an inverted landing in an experimental test. (b) Corresponding simulation image at the same point in time. (c) Schematic of the dynamic perching policy network. (d) Visual depiction of the dynamic perching sequence, illustrating the progression from flight, through the triggering ($a_{Trg}$) and rotational action ($a_{Rot}$), to the body-swing and final landing stage.}
    \label{fig:Introduction_Figure}
    \vspace*{-5mm}
\end{figure}

Works such as Thomas et. al \cite{thomas2016aggressive} and Mao et. al \cite{mao2022robust} have developed methods for landing quadrotor systems successfully onto inclined surfaces. These methods rely heavily on generating full-state trajectories that align the robot almost perfectly parallel to the landing surface upon impact, while placing less emphasis on how factors such as system geometry, perching dynamics, or robot maneuverability contribute to landing success. As these factors were implicitly accounted for and embedded within the trajectory constraints. Additionally, most research has focused on rigid landing gear attachments \cite{kim2021autonomous,yu2022implementation,hsiao2022mechanically}, with few exploring how passive flexibility contributes to landing success \cite{ni2022research,huang2021biomimetic}.

Therefore, the purpose of this work is to further the investigation of how robot (landing gear) geometry, mechanical properties and maneuverability affects landing capabilities across different robot sizes and landing surface orientations, while aiming at achieving universal dynamic perching capabilities. To this goal, this study expanded the framework developed by the author's previous work for performing Deep Reinforcement Learning (DeepRL) to explore the landing capabilities of a nano-scale quadrotor robot \cite{habas2022deep}, and using simulation with experimental validation through zero-shot Sim-to-Real transfer of the trained policy (Figure \ref{fig:Introduction_Figure}a,b).

The rest of the paper is organized as followed. Section II provides a description of our methodology and expansion of our Deep Reinforcement Learning Policy framework to universal dynamic perching. Section III details the experimental validation for the efficacy of our simulation trained landing policies and simulation behaviors on physical robot systems as well as the results of our study. Section IV then concludes the study and provides direction for future work.

\section{Methodology}

\subsection{Non-Dimensionalization of System Geometry}

\begin{figure}[!t]
    \centering
    \includegraphics[width=1.0\columnwidth]{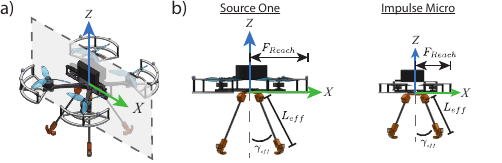}
    \caption{(a) Depiction of X-Z plane system geometry is projected onto. (b) Source One and Impulse Micro quadrotors with non-dimensionalized Semi-Narrow Short leg configuration}
    \label{fig:System_Geometry_Figure}
\end{figure}

\noindent Quadrotor systems across different scales and manufacturers often exhibit variations in their geometric proportions. To generalize robotic landing capabilities across these diverse systems, we developed a framework to non-dimensionalize the geometric relationship between a robot's leg configuration and body dimensions. This framework projects the system's geometry onto the robot's X-Z plane (Figure \ref{fig:System_Geometry_Figure}a) and defines a set of key dimensionless parameters.

The Forward Reach ($F_{Reach}$) is defined as the maximum distance from the system's vertical axis to the furthest point of the drone’s body, typically represented by the propellers or prop-guards. The effective leg length ($L_{eff}$) is defined as the distance from the robot's origin to the tip of the leg's attachment pad, projected onto the X-Z plane. The ratio of these values ($F_{Reach}/L_{eff}$), along with the angle ($\gamma$) between the vertical axis and the projected leg length, defines a set of unique dimensionless parameters for scaling across different drone sizes, as shown in Figure \ref{fig:System_Geometry_Figure}b. Assuming similar angular acceleration, we hypothesized that systems with identical dimensionless values exhibit comparable landing capabilities. To test this hypothesis, we designed, built, and modeled two custom drone systems of different sizes and landing gear designs (Table \ref{tab:Drone_Parameters}).

\begin{table}
    \centering
    \caption{Leg Design Configurations (Non-Dimensionalized)}
    \label{tab:Drone_Parameters}
    \begin{tabular}{p{2.1cm}p{1.9cm}cc}
        \hline 
        \makecell{Quadrotor Frame} & Leg Config & \makecell{Length Ratio: \\ $L_{eff} / F_{Reach}$} & \makecell{Leg Angle: \\ $\gamma$ (deg)} \\ 
        \\[-1em]
        \hline
        \\[-1em]
        \text{Source One V5}
        & \text{Wide-Long} & 1.49 & $52.4^{\circ}$ \\
        (\text{12" Diameter})
        & \text{Semi Narrow-Short} & 1.08 & $27.3^{\circ}$ \\
         \\

        \text{Impulse Micro}
        & \text{Wide-Long} & 1.47 & $53.1^{\circ}$ \\
        (\text{7" Diameter})
        & \text{Semi Narrow-Short} & 1.18 & $29.5^{\circ}$ \\
        
        \hline
    \end{tabular}
    \vspace*{-5mm}
\end{table}

\subsection{Multi-Surface Landing Policy Formulation}

\noindent To generalize the Reinforcement Learning framework established by Habas et al. 
\cite{habas2022deep} to landing on surfaces with arbitrary orientations beyond inverted, we augmented the input state vector of the policy network. It now consists of the following: $\left[ \tau, \vartheta_x, D_\perp, \theta_{plane} \right]$, which represent the system's time-to-contact, transverse velocity, distance to the landing surface, and the landing surface orientation, respectively.

The first sensory cue used as an input to our generalized landing policy is time-to-contact, $\tau$, which provides predictive information about when to initiate a perching maneuver \cite{liu2019flies,lee1993visual}. Here, $\tau$ is determined by defining a circle around the robot with a radius equal to its effective leg length ($L_{eff}$), representing all reachable orientations as the robot rotates (see Figure \ref{fig:Introduction_Figure}d). In practice, $\tau$ represents the time until the landing surface is within reach of the robot, and is computed by dividing the perpendicular distance between this circle and the landing surface ($D_\perp$) by the robot's velocity component perpendicular to the surface ($V_\perp$).

\begin{equation}
    \label{Eq:Tau_Eq}
    \tau = \frac{D_{\perp}}{V_{\perp}}.
\end{equation}

The second sensory cue is a visual observable term, $\vartheta_x$, which relates to the transverse optical flow observed by the robot and encodes information about the system's tangential velocity relative to the landing surface ($V_{\parallel}$) \cite{srinivasan2000honeybees,chahl2004landing}. This value is emulated by dividing the tangential velocity component by the distance described above,

\begin{equation}
    \label{Eq:Theta_x_Eq}
    \vartheta_x = \frac{V_{\parallel}}{D_{\perp}}.
\end{equation}

Additionally, the landing policy input vector directly includes the value $D_\perp$ and the landing surface angle ($\theta_{plane}$). As this work primarily focuses on exploring the relationship between system geometries, these values were emulated rather than obtained using physical sensors or image processing algorithms. It is important to note that the variables $\tau$ and $\vartheta_x$ can be directly estimated onboard using cameras or optical flow sensors, while $D_\perp$ and $\theta_{plane}$ can be determined through computer vision algorithms or laser distance sensors, enabling fully onboard sensory acquisition \cite{horn1981determining,horn2009hierarchical,chirarattananon2018direct}.

To facilitate the landing sequence, the robot follows a collision trajectory toward the surface with a specified speed, $\|\textbf{V}\|$, and angle relative to the horizon, $\angle \textbf{V}$. At each timestep, the state vector is input into the policy network, which outputs two actions: a triggering action, $a_{Trg}$, and a rotation action, $a_{Rot}$. When $a_{Trg}$ exceeds an arbitrary threshold, the system initiates a rotation and samples an angular acceleration ($a_{Rot}$) from the policy output. The robot then rotates until it either passes $90^\circ$ or makes contact with the landing surface, indicated by a spike in system acceleration. If contact is made, the robot transitions to a body-swing for final landing orientation. More details of this approach are provided in Habas et al \cite{habas2022deep}.

\subsection{Reward Function Design}

\noindent To optimize the control policy for emergent landing behaviors across various surface orientations and minimize bias from conventional methods, such as handcrafted policies or model predictive control, we carefully defined terms that constructed the reward function based on basic landing fundamentals and trained the policy network to thoroughly explore the landing policy-space. We chose the Soft Actor-Critic (SAC) algorithm for training due to its entropy-based exploration structure, which promotes extensive exploration of the policy-space \cite{haarnoja2018soft}. The fundamental principles we adopted to design the reward function are: 1) optimize the robot's impact orientation for efficient momentum transfer about the contact point; 2) utilize gravity during the body-swing stage; 3) ensure a landing orientation that maximizes the distance between the system’s propellers and the landing surface to avoid unnecessary collisions.

The overarching reward function was designed based on curriculum learning, where each aspect of the landing process incrementally increases the total reward, guiding the system toward mastering more complex behaviors. The reward function is a weighted sum, from the reward vector $\vec{R} = [R_{\tau_{trg}}, R_{D_{pad}}, R_{\vec{g}}, R_{\vec{L}}, R_{\phi}, R_{Legs}]$ with corresponding weights $\vec{W}_R = [0.1, 0.4, 1.0, 1.0, 2.0, 2.0]$, resulting in the scalar reward value used for training.

\begin{figure}[!t]
    \centering
    \includegraphics[width=1.0\columnwidth]{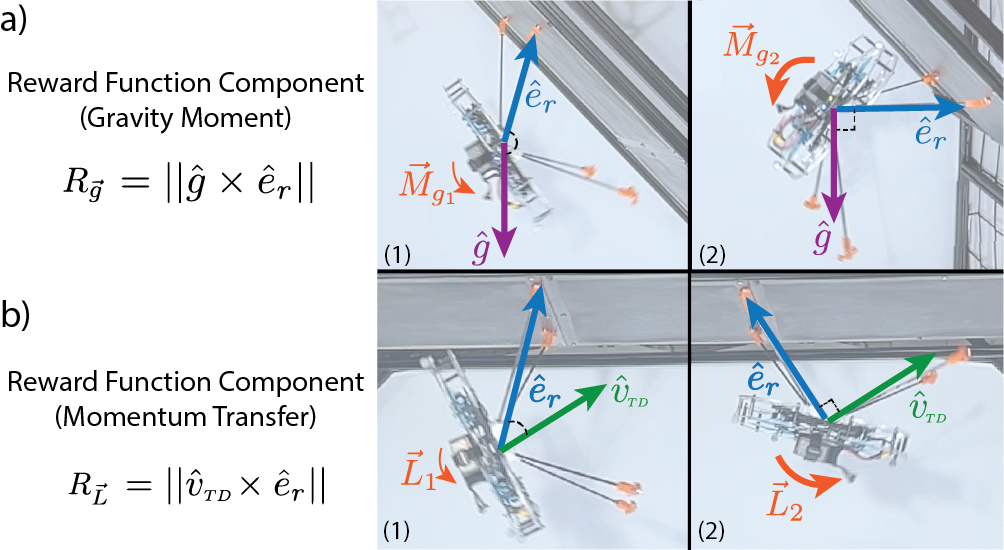}
    \caption{(a) Gravitational moment contributions: (a1) suboptimal moment; (a2) optimal moment with gravity vector perpendicular to leg vector. (b) Angular momentum transfer: (b1) suboptimal transfer; (b2) optimal transfer with touchdown velocity vector ($\hat{v}_{TD}$) perpendicular to leg vector.}
    \label{fig:MomentumGravityTransfer_Figure}
    \vspace*{-5mm}
\end{figure}

The first reward term, $R_{\tau_{trg}}$ (\ref{eq:R_Tau}), uses an exponential function to guide the robot into timing its maneuver as late as possible to learn to achieve surface contact and is based on the triggering $\tau$-value ($\tau_{trg}$) and a scaling term ($k_1$). The second term, $R_{D_{pad}}$, minimizes the distance achieved between the quadrotor’s foot-pad and the surface using $\min(D_{pad}(t))$ and scaling term $k_2$ (\ref{eq:R_D}). Next, $R_{\vec{g}}$ (\ref{eq:R_GM}) maximizes the gravitational contribution towards the body-swing by ensuring the leg vector ($\hat{e}_r$) is perpendicular to gravity ($\hat{g}$) (Figure \ref{fig:MomentumGravityTransfer_Figure}a). Similarly, $R_{\vec{L}}$ optimizes angular momentum transfer by aligning the leg vector ($\hat{e}_{r}$) perpendicular to the touchdown velocity ($\hat{v}_{TD}$), via equation (\ref{eq:R_LT}) (Figure \ref{fig:MomentumGravityTransfer_Figure}b).

The reward component $R_{\phi}$ (\ref{eq:R_Phi}) optimizes the robot's impact orientation ($\phi_{impact}$) to balance avoiding propeller contact with achieving a stable landing, using $\phi_{impact}$ and the minimum orientation where legs contact without body impact ($\phi_{min}$). The final term, $R_{Legs}$ (\ref{eq:R_legs}), maximizes the number of legs attached to the landing surface, with an additional penalty of $R_{legs} \leftarrow R_{legs} - 0.25$ for body or propeller contact to discourage damage to the quadrotor.

\begin{equation}
    \label{eq:R_Tau}
    R_{\tau_{trg}} = \begin{cases} 
    1, & \text{if } \tau_{trg} < 0 \\
    e^{-k_1(\tau_{trg})}, & \text{if } \tau_{trg} \geq 0
    \end{cases}
\end{equation}

\begin{equation}
    \label{eq:R_D}
    R_{D_{pad}} = \begin{cases} 
    1, & \text{if } \min(D_{pad}(t)) < 0 \\
    e^{-k_2(\min(D_{pad}(t)))}, & \text{if } \min(D_{pad}(t)) \geq 0
    \end{cases}
\end{equation}

\begin{equation}
    \label{eq:R_GM}
    R_{\vec{g}} = ||\hat{g} \times \hat{e}_r||
    \vspace{-4mm}
\end{equation}

\begin{equation}
    \label{eq:R_LT}
    R_{\vec{L}} = ||\hat{v}_{TD} \times \hat{e}_r||
    \vspace{-4mm}
\end{equation}

\begin{equation}
    \label{eq:R_Phi}
    R_{\phi} = \begin{cases} 
    \frac{\phi_{impact}}{\text{avg}(\phi_{min},180^\circ)}, & \text{if } \phi_{min} < |\phi_{impact}| \leq 180^\circ \\
    0.5 \cdot \frac{\phi_{impact}}{\phi_{min}}, & \text{if } 0^\circ < |\phi_{impact}| \leq \phi_{min}
    \end{cases}
\end{equation}

\begin{equation}
    \label{eq:R_legs}
    R_{Legs} = 
            \begin{cases} 
            1.0 & N_{legs} = 3 \ || \ 4 \\
            0.5 & N_{legs} = 1 \ || \ 2 \\
            0 & N_{legs} = 0 \\
            \end{cases}.
\end{equation}

The reward weights ($\vec{W}_R$) were chosen to prioritize behaviors that maximize leg contact with the landing surface, optimize impact orientation, and enhance gravity and momentum transfer effects. Components like triggering distance and timing have a stronger influence at the start of training, but their impact greatly diminishes as reliable contact is established, reducing bias in the final behavior.

\subsection{Simulation Setup and Policy Training}

\noindent To allow for training of our system in simulation and zero-shot Sim-to-Real transfer of the learned policies to our physical robots, accurate modeling was paramount. To achieve this, we utilized the Gazebo simulation environment and  simulated the inelastic attachment of the footpads and the resulting ball joint behavior which allows the robot to swing freely around the attachment point. This simulation also supports modeling complex joint behaviors such as hip joint flexibility and damping ratios.

The modeling and system identification for both the Source One and Impulse Micro quadrotor systems followed the approach by Habas and Cheng \cite{habas2024flies}. We used the bifilar pendulum method to estimate the system's rotational moment of inertia, and a thrust stand to determine motor-thrust and motor torque constants, the thrust-motor command curve, and the motors' time-constant behavior—modeled as a first-order dynamic system. This approach enabled us to establish the relationship between the necessary command signals and the resulting motor behavior for flight.

Training in our 3D simulation environment was conducted using  implemented via the StableBaselines3 RL package \cite{raffin2021stable}. Policy convergence typically occurred within 1,000 episodes or landing attempts. To maintain onboard processing efficiency, we used a small neural network with three hidden layers of 10 nodes each. During training, the landing gear configuration was fixed, while the plane angle was randomly varied among $[0^\circ, 45^\circ, 90^\circ, 135^\circ, 180^\circ]$. The robot executed a collision trajectory with a flight speed randomly sampled from the uniform distribution $\|\textbf{V}\|\sim \mathcal{U}[1.0, 5.0]$ m/s and a randomly sampled flight angle relative to the horizon ($\angle\textbf{V}$) that ensured intersection with the plane while maintaining a positive x-velocity. Following this framework, a continuous generalized policy was learned over the range of flight speeds and plane angles encountered during training.

\subsection{Experimental Setup}

\begin{figure}[!t]
    \centering
    \includegraphics[width=0.9\columnwidth]{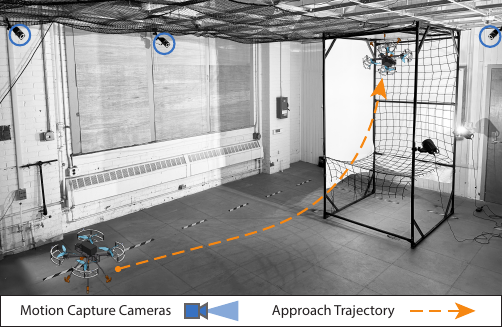}
    \caption{Experimental setup with an illustrated flight trajectory.}
    \label{fig:ExperimntalSetup_Figure}
    \vspace*{-5mm}
\end{figure}

\noindent The two quadrotor systems (Table \ref{tab:Drone_Parameters}) were built using Bitcraze Crazyflie Bolt flight controllers carbon fiber rods with 3D-printed flexible hip joints inset with spring-steel sheets of varying thickness to altar the torsional stiffness. The damping ratio of these joints were in the range of 0.3 to 0.5 and torsional stiffness values of $K = [0.4, 1.4, 8.5]$ $\frac{N \cdot m}{rad}$ were estimated using a static deflection method. 

The footpads, consisting of a custom 3D-printed compliant mechanism (made from Shore 95A hardness TPU), were designed to provide desired ball-joint behavior at attachment point (see video attachment). They were equipped with magnets for adhesion to the steel landing surface. The freely rotating behavior of the footpads accurately represents the ball-joint dynamics simulations, which model pure rotation at the contact point. Notably, the magnets’ attractive force only becomes effective within 10mm of the landing surface; beyond which the magnetic force has minimal impact during the landing sequence. The landing surface was constructed using an extruded aluminum frame to hold the steel sheets, configurable for continuous landing surface angles (Figure \ref{fig:ExperimntalSetup_Figure}). The tested angles include ceiling-type, wall-type, ground, and inclined surfaces, covering $0^\circ$, $45^\circ$, $90^\circ$, $135^\circ$, and $180^\circ$.

Communication with the robot was handled using the Crazyswarm package \cite{preiss2017crazyswarm}, which interfaces ROS messages with Bitcraze's CRTP messages for real-time data logging and command streaming. During experimental tests, state data was streamed from a Vicon motion capture system at 100 Hz and used for onboard trajectory tracking. 


For experimental testing, the 3D simulation-trained policy network for each robot and leg configuration was directly uploaded in a zero-shot manner to the physical robot. Zero-shot Sim-to-Real transfer was implemented due to the collision nature of our experiments which can damage the robots. In each experimental test, a desired flight speed ($\|\textbf{V}\|$) and flight angle ($\angle\textbf{V}$) were selected, and a flight trajectory was generated onboard to meet these approach conditions. The robot would accelerate along a collision course with the landing surface. At each timestep, the current state was fed into the trained policy network, and the outputs determined if a rotational maneuver should be initiated based on if $a_{Trg}$ exceeded a threshold. Once triggered, the rotational action, dictated by a constant angular body acceleration ($a_{Rot}$), was executed, allowing the robot to establish leg contact and body swing into the final landing position. The final number of leg contacts and the extent of body/propeller collision was determined using high-speed video footage of the landings.


\section{Results and Discussion}

\subsection{Simulation Results and Experimental Validation for Landing on Surfaces with Variable Orientation}

\begin{figure}[!t]
    \centering
    \includegraphics[width=1.0\columnwidth]{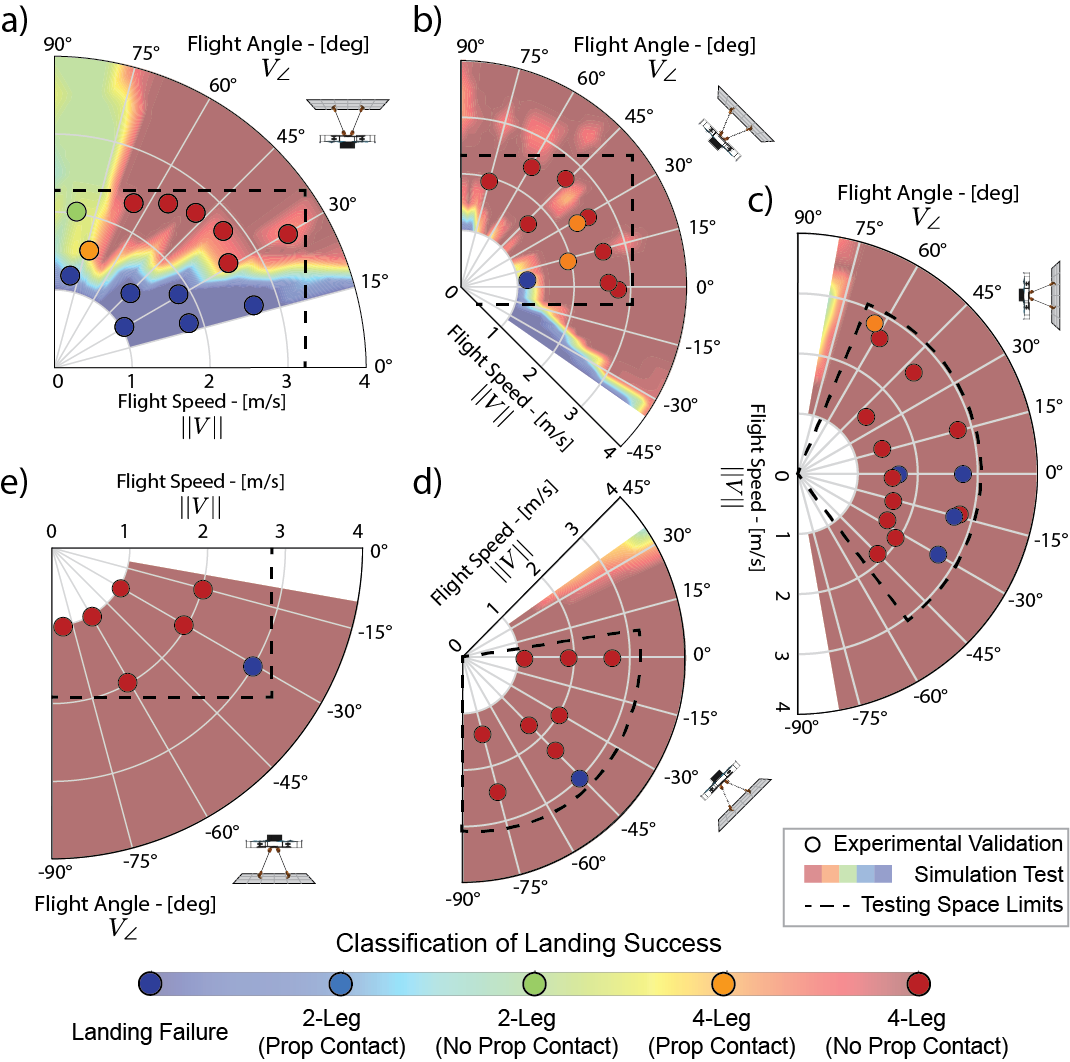}
    \caption{Polar plots illustrating the landing success rate (color spectrum) for the complete set of approach conditions tested in simulation across five surface orientations [0°, 45°, 90°, 135°, 180°], labeled a-e, and experimentally validated conditions (colored dots). The dashed box indicates the range of conditions testable within the constraints of our experimental setup (due to limited thrust capacity and lab space). Radial lines represent constant flight speeds, while angular lines denote the flight angle relative to the horizon during the quadrotor’s approach. Results are based on the Semi-Narrow Long leg configuration on the Source One quadcopter.}
    \label{fig:Experimental Validation Figure}
    \vspace*{-5mm}
\end{figure}

\noindent The policy network was successfully trained across a broad spectrum of velocity magnitudes and flight angles for five landing surface orientations (see video attachment). To assess the landing performance in simulation, the trained policy for each approach condition was tested five times to compute the average success rate, which was then smoothed for visualization (Figure \ref{fig:Experimental Validation Figure}). The results show that universal landing  success (i.e., 4-leg landing) can be achieved in all landing surface orientations for nearly all approach conditions except for the inverted surface, which was the most challenging case for landing success with a reduced set of viable approach conditions (Figure \ref{fig:Experimental Validation Figure}a).

To validate the dynamic landing policy in experimental settings, we performed a zero-shot Sim-to-Real transfer, primarily focused on the Semi-Narrow Short leg configuration with our Source One (12" diameter) quadrotor. We subjected the quadrotor to flight conditions similar to those in the simulation, taking into account the spatial and acceleration limits of our experimental setup (Figure \ref{fig:Experimental Validation Figure}). We observed that landing performance in our experiments matched closely to those in simulation for inverted landing. In the $0^\circ$, inverted (ceiling) landing scenarios (Figure \ref{fig:Experimental Validation Figure}a), the simulation shows a minimum vertical flight velocity to achieve any level of leg contact with the landing surface, and a required tangential velocity to convert two-leg hanging landings into desired four-leg landings. This distinction between two-leg and four-leg landings arises from the need for a sufficient level of momentum transfer from the robot's translational motion to angular motion \cite{habas2024flies}. 

Similar experimental validation procedures were conducted for other landing surface orientations (Figure \ref{fig:Experimental Validation Figure}b-e). The match between simulation and experimental results remained close in general, except for the wall landings (90-degree plane) with downward approach conditions with higher rate of speed (Figure\ref{fig:Experimental Validation Figure}c). This discrepancy can be attributed to the extreme angular accelerations of 90 rad/s² prioritized by our learned landing policy in simulation. High angular acceleration is beneficial for inverted landings but detrimental for wall or ground-based landings, due to the limited time window of viable impact orientations, which rendered the landing success more sensitive to increased data lag or system noise in experimental testing. This suggests that for ground or wall-based landings, lower angular accelerations should be prioritized when developing policies.

\begin{figure}[!t]
    \centering
    \includegraphics[width=1.0\columnwidth]{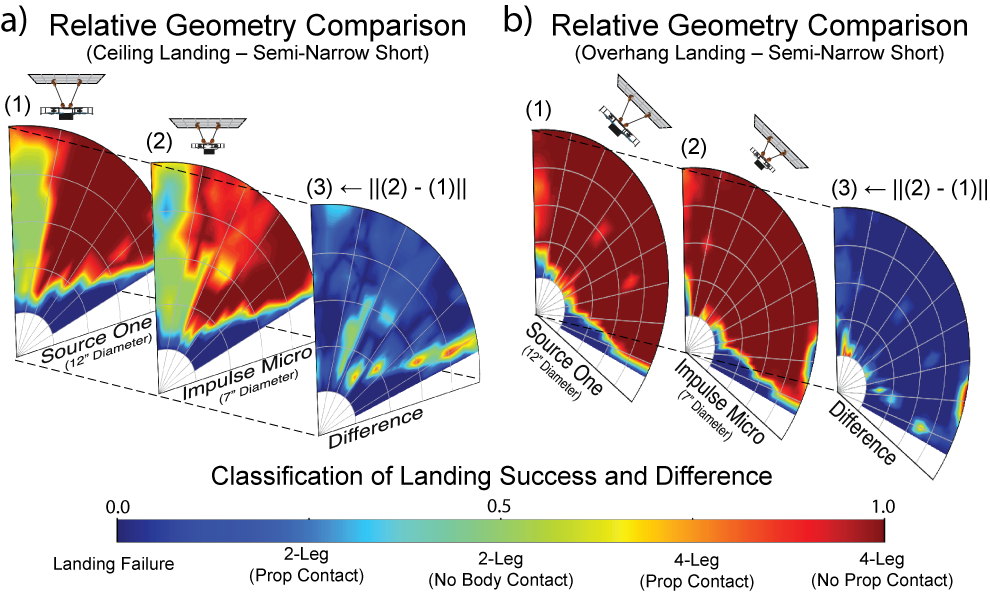}
    \caption{(a) Polar plots comparing inverted landing ($0^\circ$ - surface) capabilities of the Source One and Impulse Micro drones with similar dimensionless leg parameters; the third plot illustrates the absolute difference between them and validates our hypothesis. (b) Corresponding depiction for a $45^\circ$ overhanging landing surface.}
    \label{fig:Geometry_Comparison_Figure}
    \vspace*{-5mm}
\end{figure}

\subsection{Landing Performance Across Robot Scales and Dimensionless Geometric Parameters}

\noindent To test the hypothesis that maintaining geometric proportions across different robot scales ensures consistent perching behavior, we compared the simulation results between Source One and Impulse Micro quadrotors for landing on a $0^\circ$ ceiling (Figure \ref{fig:Geometry_Comparison_Figure}a) and a $45^\circ$ overhang surface (Figure \ref{fig:Geometry_Comparison_Figure}b). 

The comparison shows minimal differences in landing capabilities between the two robots for both cases, with only slight discrepancies appearing around the vertical velocity threshold (Figure \ref{fig:Geometry_Comparison_Figure}a), which can be attributed to the stochastic nature of the policy. The $90^\circ$ wall, $135^\circ$ incline, and $180^\circ$ ground surfaces were omitted for brevity as they yield similar conclusions of success for all conditions; like those shown in Figure \ref{fig:Experimental Validation Figure}c-e. 
 Together, the above results supports our hypothesis, which underscores the relationship between body geometry and leg geometry in determining landing success capabilities across robot sizes.

\subsection{Effects of Leg Stiffness and Damping}

\begin{figure}[!t]
    \centering
    \includegraphics[width=1.0\columnwidth]{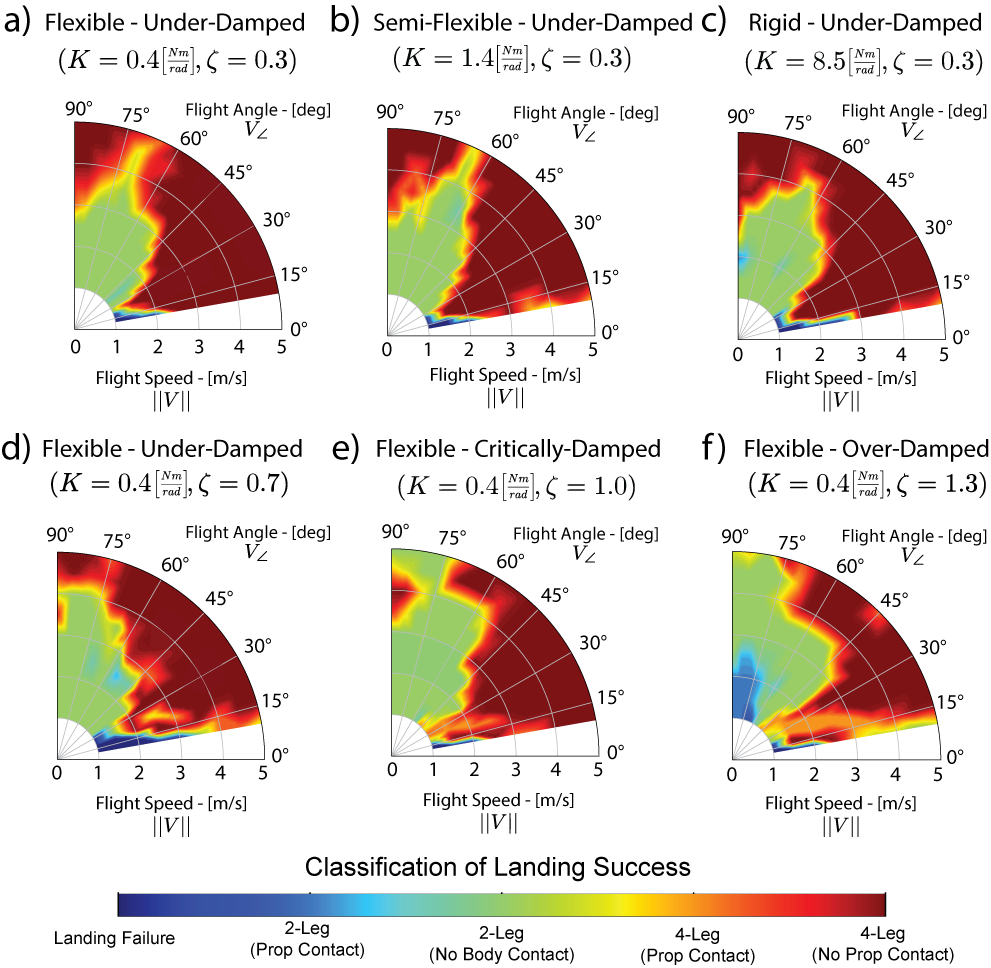}
    \caption{(a-c) Polar plots illustrating the landing performance of the Impulse Micro with a Wide-Long leg configuration, showing inverted landing scenarios across different hip stiffness values. (a,d-f) Corresponding plots depicting the effects of varying hinge damping ratios.}
    \label{fig:Stiffness_Comparison}
    \vspace*{-6mm}
\end{figure}

\noindent We evaluated the effects of stiffness and damping of the robot hip joint using the Impulse Micro quadrotor robot with Wide-Long leg configuration for inverted landing (Figure \ref{fig:Stiffness_Comparison}). Note that landing on other surface orientations was largely successful over all approach conditions and therefore not ideal to reveal the effects of leg stiffness and damping. These policies were trained in simulation with varied leg stiffness, followed by experimental validation (see video attachment).

As shown in Figure \ref{fig:Stiffness_Comparison}a-c, the landing performance remained comparable for the hip stiffness ranging from extremely flexible to semi-flexible, and finally to effectively rigid (while maintaining an underdamped behavior). On the other hand, the landing performance with varied hip joint damping ratios ranging from under-damped to critically-damped and over-damped (where the joint stiffness remains flexible) showed considerable difference for inverted landing with near vertical approach conditions, while remaining consistent in other conditions(Figure \ref{fig:Stiffness_Comparison}a,d-f). In  these vertical approach conditions, the robot achieved only two-leg (hanging) landings on the ceiling. This outcome is primarily due to the overly high damping in the hinge joint, which dissipates energy too quickly at touchdown, preventing the robot from leveraging the momentum transfer from vertical translation to rotation. In addition, limited changes are observed under higher forward velocity conditions. This behavior stems from a shift in landing dynamics, where the robot leverages its forward momentum to swing into the final landing position, a scenario involving minimal flex at the hip joint.

These findings suggest that landing success in inverted cases heavily depends on the efficient transfer and retention of momentum. Notably, a substantial increase in system damping diminishes landing performance, while variations in joint stiffness have minimal effect.

\subsection{Angular Acceleration Limits and Velocity Threshold}

\begin{figure}[!t]
    \centering
    \includegraphics[width=1.0\columnwidth]{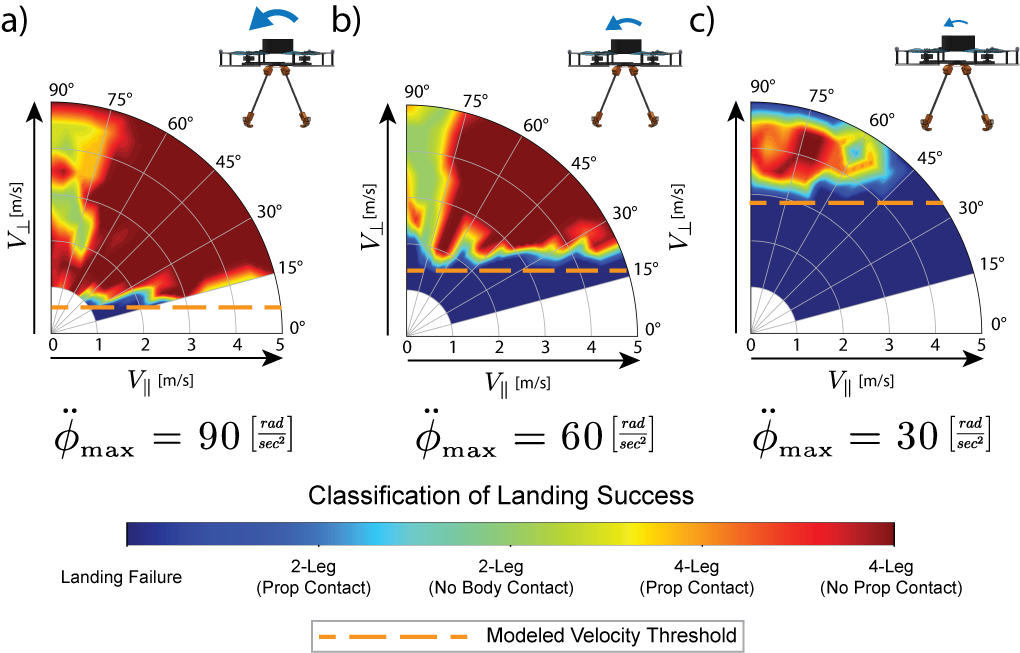}
    \caption{Landing performance of our Source One robot with Semi-Narrow Short leg configuration and performing inverted landings with variations on the angular acceleration limit.}
    \label{fig:Angular_Acceleration_Figure}
    \vspace*{-6mm}
\end{figure}

\noindent A notable result from this work is the identification of a perpendicular velocity threshold that depends on the landing surface orientation, the leg configuration of the robot, and the robot's maneuverability (angular acceleration limit). As the angular acceleration capabilities decrease, the required perpendicular velocity threshold increases, with any approach conditions below this threshold resulting in landing  failure. Figure \ref{fig:Angular_Acceleration_Figure} illustrates this relationship, showing the landing performance for different angular acceleration capabilities of the Semi-Narrow configuration on our Source One Drone during landings on an (inverted) ceiling-type surface.

This phenomenon stems from the system's kinematics. As the robot translates and rotates during the rotation maneuver, the forward propellers and foot-pads trace distinct trajectories. Along these trajectories, the minimum distance achieved between the foot-pad and the landing surface ($\min(D_{pad}(t))$) must be closer to the landing surface than the minimum distance from the propeller to the surface ($\min(D_{prop}(t))$). In the inverted landing scenario, if the system rotates too slowly or lacks sufficient vertical velocity, the robot may fall away from the surface before its legs rotate into a favorable inverted orientation. However, with enough angular acceleration and velocity, the system can time its rotation so the legs are oriented toward the surface just as the robot reaches its peak height. This relationship suggests a predictive model for the minimum velocity conditions required for robots, with any landing geometry, to land on surfaces of any orientation. Our simplified kinematic modeling of this predicted perpendicular velocity, represented by the dashed line in Figure \ref{fig:Angular_Acceleration_Figure}, aligns closely with the transition from landing failures to successful touchdowns.

\section{Conclusion and Future Work}

\noindent By expanding upon previous work using a Deep Reinforcement Learning framework by Habas et al. \cite{habas2022deep}, we successfully demonstrated universal dynamic perching capability on surfaces of variable orientation with robots of variable sizes. We also developed and applied a non-dimensionalization framework for assessing the landing capabilities of quadrotors across different surface orientations and sizes, while investigating the contributions of flexible hip joints and system maneuverability on perching success.

Our findings confirmed that maintaining geometric proportions results in similar perching behaviors across robot sizes, with our simulation effectively matching real-world dynamics across varied landing scenarios. Our investigation into landing gear stiffness and damping revealed minimal impact from changes in joint stiffness, whereas variations in damping ratios significantly influenced landing outcomes but only under certain approach conditions. Additionally, this study highlights key relationships between system kinematics and robot geometry, such as the inverse relationship between angular acceleration and the velocity threshold required for successful landings (i.e., as angular acceleration decreases, the velocity threshold increases).

In future work, we aim to refine the relationship between mechanical damping, system geometry, and angular acceleration to develop more predictive, robust models for autonomous quadrotor perching. A custom-designed miniature controllable landing gear will also be developed for landing on surfaces commonly found in urban environments. We will also explore integrating onboard sensors and real-time data processing to enhance quadrotor autonomy and situational responsiveness, broadening the potential applications of these technologies in complex operational environments.






\bibliography{IEEEabrv, refs}

\begin{thebibliography}{10}
\providecommand{\url}[1]{#1}
\csname url@samestyle\endcsname
\providecommand{\newblock}{\relax}
\providecommand{\bibinfo}[2]{#2}
\providecommand{\BIBentrySTDinterwordspacing}{\spaceskip=0pt\relax}
\providecommand{\BIBentryALTinterwordstretchfactor}{4}
\providecommand{\BIBentryALTinterwordspacing}{\spaceskip=\fontdimen2\font plus
\BIBentryALTinterwordstretchfactor\fontdimen3\font minus \fontdimen4\font\relax}
\providecommand{\BIBforeignlanguage}[2]{{%
\expandafter\ifx\csname l@#1\endcsname\relax
\typeout{** WARNING: IEEEtran.bst: No hyphenation pattern has been}%
\typeout{** loaded for the language `#1'. Using the pattern for}%
\typeout{** the default language instead.}%
\else
\language=\csname l@#1\endcsname
\fi
#2}}
\providecommand{\BIBdecl}{\relax}
\BIBdecl

\bibitem{lee1993visual}
D.~N. Lee, M.~N. Davies, P.~R. Green, and F.~.~Van Der~Weel, ``Visual control of velocity of approach by pigeons when landing,'' \emph{Journal of experimental biology}, vol. 180, no.~1, pp. 85--104, 1993.

\bibitem{carruthers2010mechanics}
A.~C. Carruthers, A.~L. Thomas, S.~M. Walker, and G.~K. Taylor, ``Mechanics and aerodynamics of perching manoeuvres in a large bird of prey,'' \emph{The Aeronautical Journal}, vol. 114, no. 1161, pp. 673--680, 2010.

\bibitem{srinivasan2000honeybees}
M.~V. Srinivasan, S.-W. Zhang, J.~S. Chahl, E.~Barth, and S.~Venkatesh, ``How honeybees make grazing landings on flat surfaces,'' \emph{Biological cybernetics}, vol.~83, no.~3, pp. 171--183, 2000.

\bibitem{liu2019flies}
P.~Liu, S.~P. Sane, J.-M. Mongeau, J.~Zhao, and B.~Cheng, ``Flies land upside down on a ceiling using rapid visually mediated rotational maneuvers,'' \emph{Science advances}, vol.~5, no.~10, p. eaax1877, 2019.

\bibitem{riskin2009bats}
D.~K. Riskin, J.~W. Bahlman, T.~Y. Hubel, J.~M. Ratcliffe, T.~H. Kunz, and S.~M. Swartz, ``Bats go head-under-heels: the biomechanics of landing on a ceiling,'' \emph{Journal of Experimental Biology}, vol. 212, no.~7, pp. 945--953, 2009.

\bibitem{zufferey2008bio}
J.-C. Zufferey, \emph{Bio-inspired flying robots: experimental synthesis of autonomous indoor flyers}.\hskip 1em plus 0.5em minus 0.4em\relax Epfl Press, 2008.

\bibitem{seo2018drone}
J.~Seo, L.~Duque, and J.~Wacker, ``Drone-enabled bridge inspection methodology and application,'' \emph{Automation in Construction}, vol.~94, pp. 112--126, 2018.

\bibitem{irizarry2012usability}
J.~Irizarry, M.~Gheisari, and B.~N. Walker, ``Usability assessment of drone technology as safety inspection tools,'' \emph{Journal of Information Technology in Construction (ITcon)}, vol.~17, no.~12, pp. 194--212, 2012.

\bibitem{mishra2020drone}
B.~Mishra, D.~Garg, P.~Narang, and V.~Mishra, ``Drone-surveillance for search and rescue in natural disaster,'' \emph{Computer Communications}, vol. 156, pp. 1--10, 2020.

\bibitem{kim2018drone}
S.~J. Kim and G.~J. Lim, ``Drone-aided border surveillance with an electrification line battery charging system,'' \emph{Journal of Intelligent \& Robotic Systems}, vol.~92, no.~3, pp. 657--670, 2018.

\bibitem{cesare2015multi}
K.~Cesare, R.~Skeele, S.-H. Yoo, Y.~Zhang, and G.~Hollinger, ``Multi-uav exploration with limited communication and battery,'' in \emph{2015 IEEE international conference on robotics and automation (ICRA)}.\hskip 1em plus 0.5em minus 0.4em\relax IEEE, 2015, pp. 2230--2235.

\bibitem{khalil2019using}
M.~I. Khalil, ``Using battery-saving modes in quadcopter relay networks,'' \emph{Physical Communication}, vol.~35, p. 100692, 2019.

\bibitem{habas2022optimal}
B.~Habas, B.~AlAttar, B.~Davis, J.~W. Langelaan, and B.~Cheng, ``Optimal inverted landing in a small aerial robot with varied approach velocities and landing gear designs,'' in \emph{2022 International Conference on Robotics and Automation (ICRA)}.\hskip 1em plus 0.5em minus 0.4em\relax IEEE, 2022, pp. 2003--2009.

\bibitem{habas2022deep}
B.~Habas, J.~W. Langelaan, and B.~Cheng, ``Inverted landing in a small aerial robot via deep reinforcement learning for triggering and control of rotational maneuvers,'' \emph{arXiv preprint arXiv:2209.11043}, 2022.

\bibitem{habas2024flies}
B.~Habas and B.~Cheng, ``From flies to robots: Inverted landing in small quadcopters with dynamic perching,'' \emph{arXiv preprint arXiv:2403.00128}, 2024.

\bibitem{thomas2016aggressive}
J.~Thomas, M.~Pope, G.~Loianno, E.~W. Hawkes, M.~A. Estrada, H.~Jiang, M.~R. Cutkosky, and V.~Kumar, ``Aggressive flight with quadrotors for perching on inclined surfaces,'' \emph{Journal of Mechanisms and Robotics}, vol.~8, no.~5, p. 051007, 2016.

\bibitem{mao2022robust}
J.~Mao, S.~Nogar, C.~Kroninger, and G.~Loianno, ``Robust active visual perching with quadrotors on inclined surfaces,'' \emph{arXiv preprint arXiv:2204.02458}, 2022.

\bibitem{kim2021autonomous}
J.~Kim, M.~C. Lesak, D.~Taylor, D.~J. Gonzalez, and C.~M. Korpela, ``Autonomous quadrotor landing on inclined surfaces using perception-guided active asymmetric skids,'' \emph{IEEE Robotics and Automation Letters}, vol.~6, no.~4, pp. 7877--7877, 2021.

\bibitem{yu2022implementation}
P.~Yu and K.~Wong, ``An implementation framework for vision-based bat-like inverted perching with bi-directionalthrust quadrotor,'' \emph{International Journal of Micro Air Vehicles}, vol.~14, p. 17568293211073672, 2022.

\bibitem{hsiao2022mechanically}
H.~Hsiao, J.~Sun, H.~Zhang, and J.~Zhao, ``A mechanically intelligent and passive gripper for aerial perching and grasping,'' \emph{IEEE/ASME Transactions on Mechatronics}, 2022.

\bibitem{ni2022research}
X.~Ni, Q.~Yin, X.~Wei, P.~Zhong, and H.~Nie, ``Research on landing stability of four-legged adaptive landing gear for multirotor uavs,'' \emph{Aerospace}, vol.~9, no.~12, p. 776, 2022.

\bibitem{huang2021biomimetic}
Z.~Huang, S.~Li, J.~Jiang, Y.~Wu, L.~Yang, and Y.~Zhang, ``Biomimetic flip-and-flap strategy of flying objects for perching on inclined surfaces,'' \emph{IEEE Robotics and Automation Letters}, vol.~6, no.~3, pp. 5199--5206, 2021.

\bibitem{chahl2004landing}
J.~S. Chahl, M.~V. Srinivasan, and S.-W. Zhang, ``Landing strategies in honeybees and applications to uninhabited airborne vehicles,'' \emph{The International Journal of Robotics Research}, vol.~23, no.~2, pp. 101--110, 2004.

\bibitem{horn1981determining}
B.~K. Horn and B.~G. Schunck, ``Determining optical flow,'' \emph{Artificial intelligence}, vol.~17, no. 1-3, pp. 185--203, 1981.

\bibitem{horn2009hierarchical}
B.~K. Horn, Y.~Fang, and I.~Masaki, ``Hierarchical framework for direct gradient-based time-to-contact estimation,'' in \emph{2009 IEEE Intelligent Vehicles Symposium}.\hskip 1em plus 0.5em minus 0.4em\relax IEEE, 2009, pp. 1394--1400.

\bibitem{chirarattananon2018direct}
P.~Chirarattananon, ``A direct optic flow-based strategy for inverse flight altitude estimation with monocular vision and imu measurements,'' \emph{Bioinspiration \& biomimetics}, vol.~13, no.~3, p. 036004, 2018.

\bibitem{haarnoja2018soft}
T.~Haarnoja, A.~Zhou, P.~Abbeel, and S.~Levine, ``Soft actor-critic: Off-policy maximum entropy deep reinforcement learning with a stochastic actor,'' in \emph{International conference on machine learning}.\hskip 1em plus 0.5em minus 0.4em\relax PMLR, 2018, pp. 1861--1870.

\bibitem{raffin2021stable}
A.~Raffin, A.~Hill, A.~Gleave, A.~Kanervisto, M.~Ernestus, and N.~Dormann, ``Stable-baselines3: Reliable reinforcement learning implementations,'' \emph{Journal of Machine Learning Research}, 2021.

\bibitem{preiss2017crazyswarm}
J.~A. Preiss, W.~Honig, G.~S. Sukhatme, and N.~Ayanian, ``Crazyswarm: A large nano-quadcopter swarm,'' in \emph{2017 IEEE International Conference on Robotics and Automation (ICRA)}.\hskip 1em plus 0.5em minus 0.4em\relax IEEE, 2017, pp. 3299--3304.

\end{thebibliography}
\bibliographystyle{IEEEtran}

\end{document}